\documentclass{article}

\usepackage{microtype}
\usepackage{graphicx}
\usepackage{subcaption}
\usepackage{booktabs}
\usepackage{hyperref}
\usepackage{multirow}

\usepackage[accepted]{icml2026}

\usepackage{amsmath}
\usepackage{amssymb}
\usepackage{mathtools}
\usepackage{amsthm}

\usepackage[capitalize,noabbrev]{cleveref}

\icmltitlerunning{VFUSE: Virulent Feature Understanding with Sparse autoEncoders}

\begin{document}

\twocolumn[
  \icmltitle{VFUSE: Virulent Feature Understanding with Sparse autoEncoders
}

  \icmlsetsymbol{equal}{*}

  \begin{icmlauthorlist}
    \icmlauthor{Michael Yu}{raft}
     \icmlauthor{Matthew L. Olson}{independent}
  \end{icmlauthorlist}

  \icmlaffiliation{independent}{Independent}
  \icmlaffiliation{raft}{Raft Bioworks}

  \icmlcorrespondingauthor{Michael Yu}{michael@raft.bio}

  \icmlkeywords{sparse autoencoders, mechanistic interpretability, protein diffusion, biosecurity, RFdiffusion}

  \vskip 0.3in
]

\printAffiliationsAndNotice{} 

\begin{abstract}
Generative models have shown remarkable progress in a variety of domains such as protein design, but such power enables the opaque generation of hazardous proteins. In this work, we introduce VFUSE (Virulent Feature Understanding with Sparse autoEncoders), a mechanistic interpretability approach that trains SAEs on diffusion-transformer activations to audit protein models for hazard-aware features. We apply VFUSE to RoseTTAFold3 and RFDiffusion3, popular open-weight models for protein folding and synthesis. We find that for certain blocks, linear probes detect hazardous designs significantly better when fit in the SAE latent space over the original model's representations: improving interpretability without sacrificing model performance. Furthermore, we identify monosemantic features from the SAE that fire only on hazardous designs at up to AUROC $0.84$ ($q < 10^{-13}$). To our knowledge this is the first SAE trained on an all-atom diffusion model and the first feature-level virulence audit of a protein design model, paving the way towards safe and interpretable protein design.
\end{abstract}

\section{Introduction}
\label{sec:intro}

Protein diffusion models can now design binders, enzymes, and oligomers that are validated to fold in the wet lab \cite{watson2023rfdiffusion,krishna2024rfaa,butcher2025rfdiffusion3,woolfson2021brief}. This capability is dual-use: given a hazardous template such as a neurotoxin active site, a diffusion model can scaffold new proteins around it. Existing biosecurity screens rely on sequence homology or sequence-level classifiers \cite{garg2008virulentpred,singh2024vfpred,sun2024dtvf,chen2025virulenthunter}. They flag a sequence after the fact but cannot explain which structural features triggered the flag, and they cannot intervene during sampling.

Mechanistic interpretability has made substantial progress in understanding language models \cite{bricken2023monosemanticity,cunningham2024sparse,templeton2024scaling} and, more recently, vision models \cite{joseph2025prisma,simonyan2014deep,Ben_Melech_Stan_2024_CVPR}. Sparse autoencoders (SAEs) decompose dense latent activations into sparse, approximately monosemantic directions, and have been applied to protein language models \cite{simon2025interplm,adams2025mechanistic,parsan2025interpretable}. But all-atom diffusion models have not been similarly probed.

\begin{figure*}[tb]
\centering
\begin{subfigure}[t]{0.245\linewidth}
  \includegraphics[width=\linewidth]{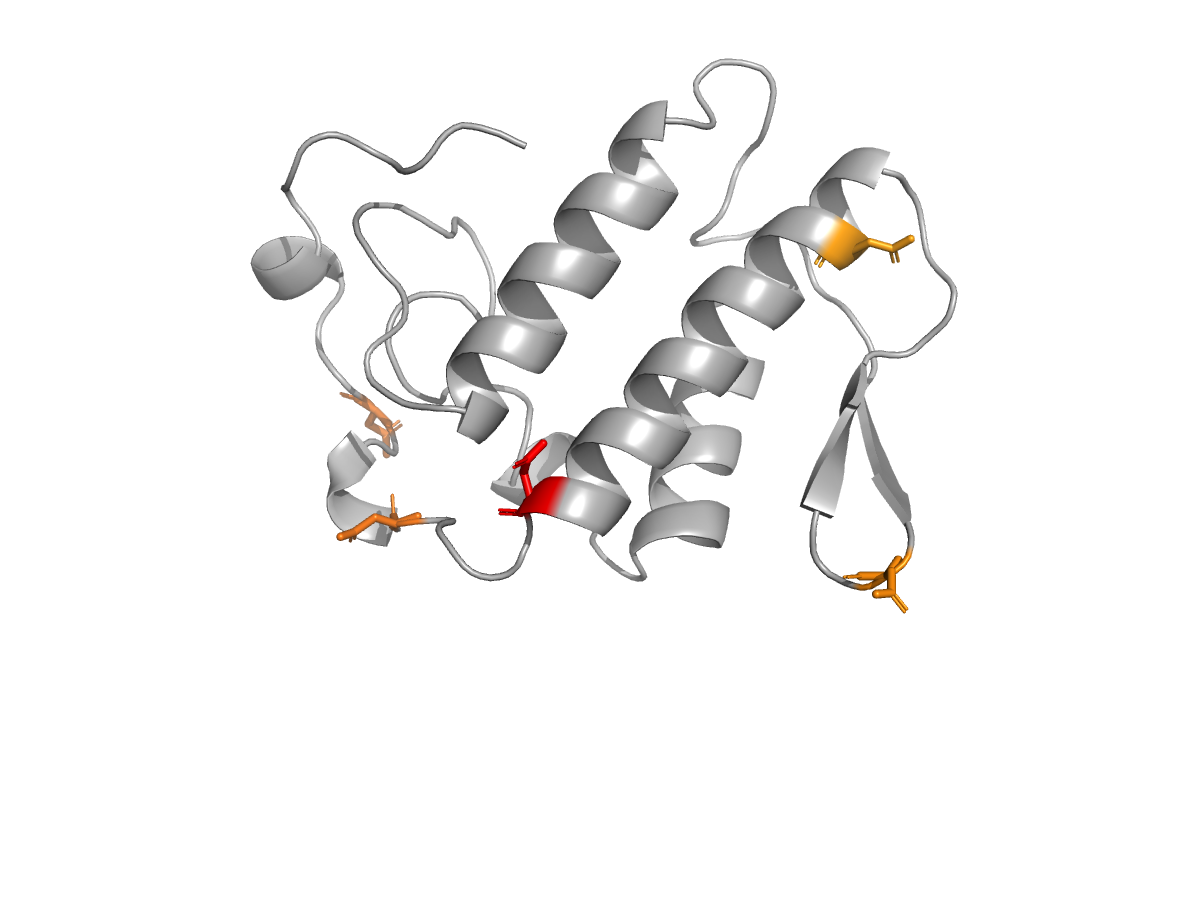}
  \caption{\textbf{Feature 639} (AUROC 0.819)\\Ammodytoxin A (P00626)\\Viper neurotoxin}
\end{subfigure}\hfill
\begin{subfigure}[t]{0.245\linewidth}
  \includegraphics[width=\linewidth]{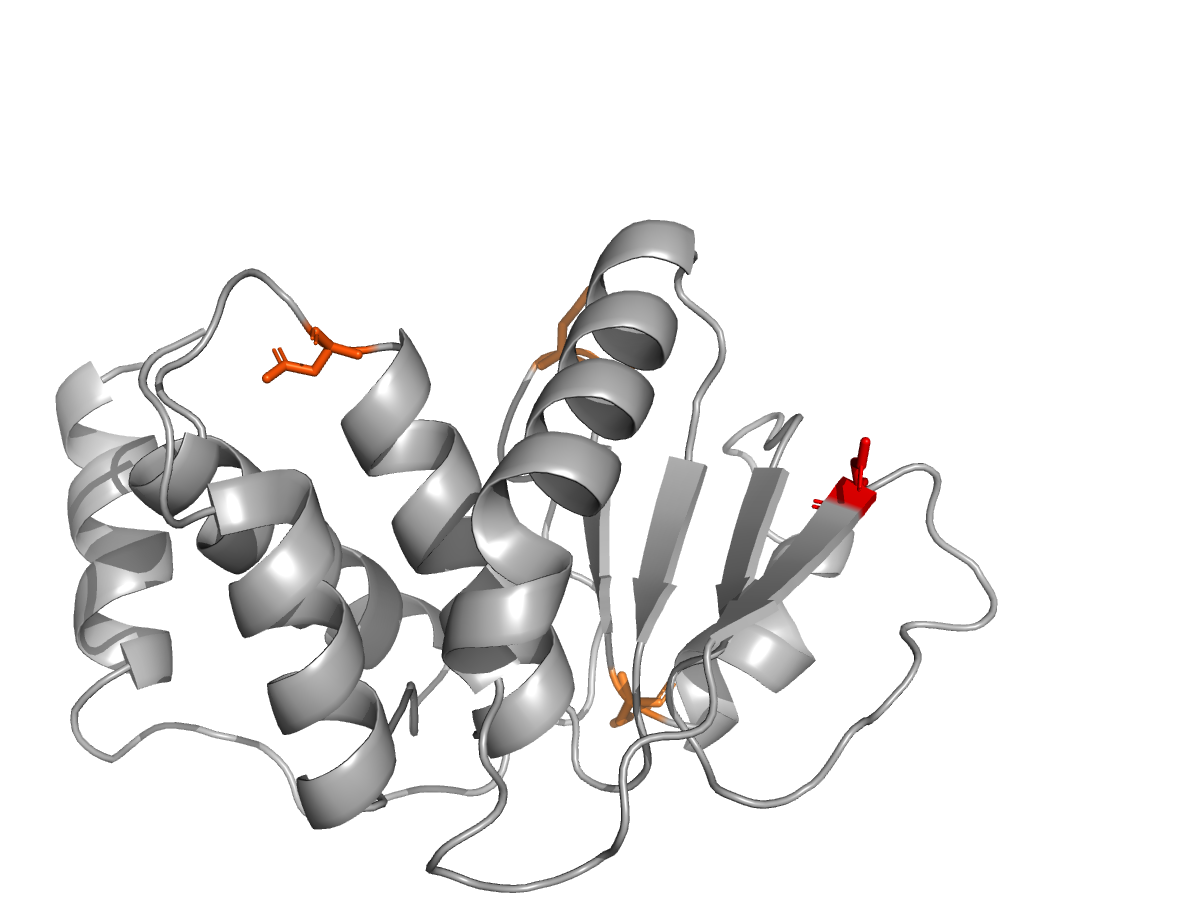}
  \caption{\textbf{Feature 60} (AUROC 0.813)\\OPG106 (A0A7H0DN78)\\Monkeypox immune evasion}
\end{subfigure}\hfill
\begin{subfigure}[t]{0.245\linewidth}
  \includegraphics[width=\linewidth]{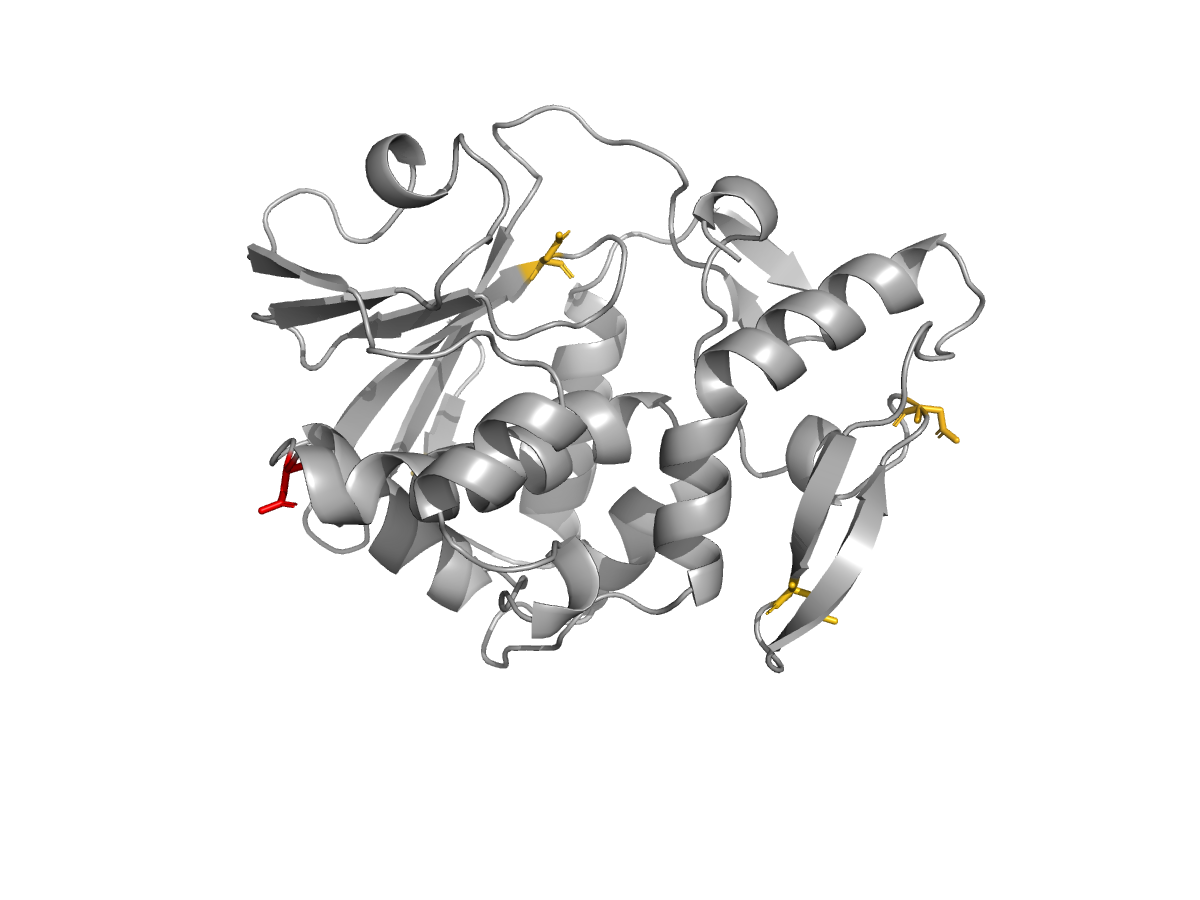}
  \caption{\textbf{Feature 170} (AUROC 0.741)\\Cucurmosin (D0VWS7)\\Ribosome-inactivating protein}
\end{subfigure}\hfill
\begin{subfigure}[t]{0.245\linewidth}
  \includegraphics[width=\linewidth]{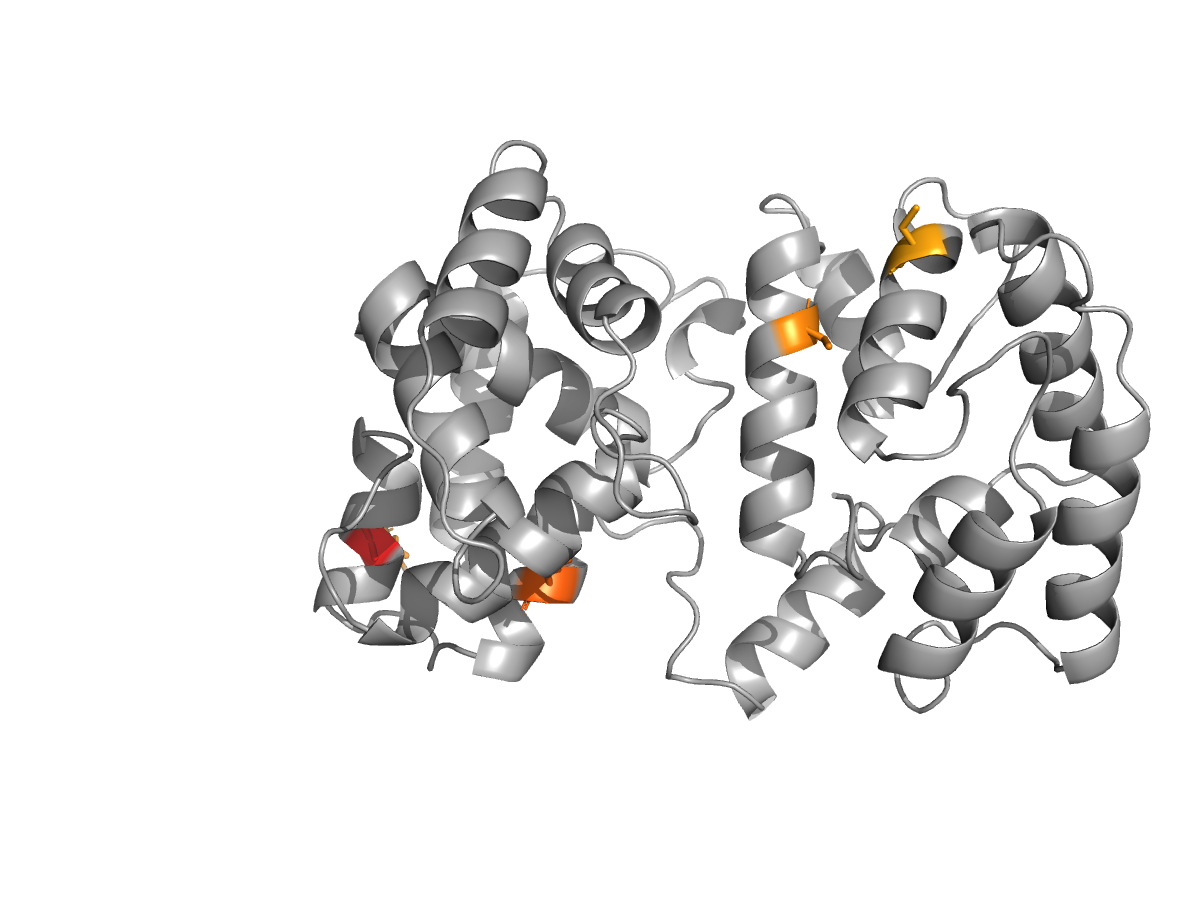}
  \caption{\textbf{Feature 351} (AUROC 0.731)\\D7L1 (A0A1S4K3K8)\\Mosquito salivary toxin}
\end{subfigure}
\caption{Top-firing residues for four leading RFD3 block-12 SAE features on partial-diffusion output structures. Red = highest activation, orange = moderate, gray = inactive. Feature 639's top-3 residues (99, 104, 109) cluster in a single alpha-helix of a viper neurotoxin. The other features fire on proteins from three other hazard classes, with similarly focal but less structurally concentrated activation.}
\label{fig:features}
\end{figure*}

Applying VFUSE to RFdiffusion3 (RFD3) and RoseTTAFold3 (RF3), we find that at block 12 of RFD3, SAE-encoded representations outperform raw activations as probe inputs, and that individual features fire selectively on hazardous designs with per-residue localization (See \Cref{fig:features}). We release SAE checkpoints at three depths in RFD3 and two in RF3, together with a catalog of hazard-associated features.

\section{Background}

\paragraph{Sparse Autoencoders}
\label{sec:background}
Given an activation vector $x \in \mathbb{R}^{d}$, an SAE is an encoder
with a nonlinear activation function $\sigma$ and a linear decoder,
\begin{equation}
    z = \sigma(W_e x + b_e), \qquad \hat{x} = W_d z + b_d,
\end{equation}
trained to minimize a reconstruction loss, optionally with a sparsity penalty:
\begin{equation}
    \mathcal{L} = \mathbb{E}_x \big[\|x - \hat{x}\|_2^2\big] + \lambda \, \mathcal{S}(z).
\end{equation}
$W_e \in \mathbb{R}^{m \times d}$, $W_d \in \mathbb{R}^{d \times m}$ with $m \gg d$. Decoder columns are constrained to unit $L_2$ norm \cite{bricken2023monosemanticity}.

\paragraph{BatchTopK.} Rather than enforcing sparsity through an $L_1$ penalty on $z$, BatchTopK \cite{bussmann2024batchtopk} bakes it into the activation function: $\sigma$ keeps only the $BK$ largest pre-activations across an entire batch of $B$ samples,
\begin{equation}
    \sigma(u) = u \odot \mathbf{1}\!\big[u \geq \tau_{B,K}(U)\big],
\end{equation}
where $\tau_{B,K}(U)$ is the $(BK)$-th largest entry of the pre-activation tensor $U \in \mathbb{R}^{B \times m}$, yielding an average of $K$ active latents per sample. At inference, $\tau$ is replaced by a learned scalar threshold so that activations are batch-independent.

\paragraph{Matryoshka.} A Matryoshka SAE \cite{bussmann2025matryoshka} fixes a sequence of nested prefix sizes $m_1 < m_2 < \cdots < m_p = m$ and, for each prefix, decodes a reconstruction $\hat{x}^{(i)}$ using only the first $m_i$ latents (and the corresponding decoder columns). The total loss
\begin{equation}
    \mathcal{L}_{\text{matryoshka}} = \sum_{i=1}^{p} w_i \, \big\|x - \hat{x}^{(i)}\big\|_2^2
\end{equation}
forces earlier (smaller) prefixes to reconstruct $x$ on their own, encouraging them to capture coarse, general-purpose features, while later prefixes refine the reconstruction with finer-grained detail.

\subsection{Related Work}
\label{sec:related}

\paragraph{SAEs for protein models.}
\citet{simon2025interplm} recover thousands of biologically interpretable features from ESM-2 \cite{lin2023evolutionaryscale}; \citet{adams2025mechanistic} link ESM-2 SAE features to Gene Ontology (GO) terms and downstream properties; \citet{parsan2025interpretable} demonstrate causal steering of ESMFold's predicted surface area. FoldSAE \cite{zarzecki2025foldsae} trains SAEs on the token-level RFdiffusion \cite{watson2023rfdiffusion} and finds secondary-structure features. However, RFD3 operates on a mixed token-and-atom representation \cite{krishna2024rfaa,abramson2024alphafold3}.

\paragraph{Virulence prediction.}
Sequence-only classifiers have evolved from SVMs \cite{garg2008virulentpred} to PLM fine-tunes \cite{sun2024dtvf,chen2025virulenthunter}. The SOTA on the standard 576/576 benchmark is DTVF \cite{sun2024dtvf} at AUROC 0.92. None provide per-residue rationale.

\section{Methods}
\label{sec:methods}

\subsection{Models and activation collection}

RFD3 \cite{butcher2025rfdiffusion3} is an 18-block all-atom diffusion model; RF3 \cite{krishna2024rfaa} is its folding counterpart. Both operate on per-residue and per-atom representations. We attach PyTorch hooks to the diffusion transformer of each model and capture activations ($d = 768$) at three blocks in RFD3 (6, 8, 12) and two in RF3 (12, 16).

\subsection{Dataset}

We draw virulent sequences from SafeProtein \cite{fan2025safeprotein} and length-matched benign sequences from UniProt \cite{uniprot2023} (excluding toxin, virulence, and viral keywords; lengths 100--300\,aa), giving $n = 275$ pairs. For RF3 only we also use ToxinPred3 \cite{rathore2024toxinpred} ($n = 1200$), but omit RFD3 due to the sequence-only nature of the dataset and lack of protein structures. For RFD3 inputs, we apply partial diffusion at $\partial t = 5$ (${\approx}5$\,\AA\ noise) to each sequence's PDB structure, simulating the activations the model produces when designing around that template. For RF3 we fold directly.

\subsection{SAE training}

For each (model, block) pair we train a Matryoshka BatchTopK SAE with dictionary size $m = 12{,}288$ ($16\times$ expansion), $K = 80$, five nested group fractions, learning rate $2.3 \times 10^{-4}$, and 20{,}000 steps on an L40 GPU, using the \texttt{dictionary\_learning} codebase \cite{marks2024dictionarylearning}. Training activations come from a held-aside corpus disjoint from SafeProtein. Trained SAEs explain $96.9\%$ of held-out activation variance ($L_0 = 79.9$, cossim $= 0.983$), with all features alive. Full hyperparameters are in \cref{app:hparams}.

\subsection{Virulence probes}

We mean-pool per-token SAE activations $z$ and raw activations $x$ over residues and denoising steps to obtain per-design vectors, then fit $L_2$-regularized logistic regression probes. We evaluate with 5-fold cross-validation under two regimes: \textit{random} (stratified) and \textit{homology-clustered}. Homology refers to shared evolutionary ancestry between protein sequences; proteins with high sequence identity tend to share structure and function. Without controlling for it, a probe trained on one family member can memorize fold-class identity rather than learning a virulence-specific signal, and it will appear to generalize when tested on a homologous sequence from the same family. We cluster all sequences with mmseqs2 at 30\% identity \cite{steinegger2017mmseqs2} and assign whole clusters to folds so that no near-homolog appears in both train and test. We also calculate a length-only baseline, resulting in below AUROC 0.55, ruling out a trivial signal.

\subsection{Univariate feature scoring}

To identify individual SAE directions associated with virulence, we compute the univariate AUROC of each feature's mean activation against the class label. We then run Mann--Whitney U tests \cite{mann1947whitney} and apply Benjamini--Hochberg FDR correction \cite{benjamini1995controlling} to control for testing all 12{,}288 features simultaneously, reporting features with $q < 0.05$. Top features are visualized in PyMOL by mapping per-residue activations onto the diffusion output structure.

\begin{table}[tb]
  \caption{Held-out AUROC (mean $\pm$ std, 5 folds). \textbf{Cluster} = homology-clustered splits; \textbf{Random} = stratified random splits. \textbf{Bold} = best per dataset. Probe: \textit{raw} = raw activations; \textit{sae} = SAE-encoded features.}
  \label{tab:probes}
  \centering
  \begin{small}
    \begin{tabular}{llrccc}
      \toprule
      Model & Block & Probe & Cluster & Random \\
      \midrule
      \multicolumn{5}{l}{\textit{SafeProtein ($n$=275)}} \\
      \cmidrule(lr){1-5}
      \multirow{6}{*}{RFD3}
        & 6  & raw & $0.592 \pm 0.027$ & $0.722 \pm 0.073$ \\
        & 6  & sae & $0.599 \pm 0.106$ & $0.717 \pm 0.089$ \\
        & 8  & raw & $0.718 \pm 0.108$ & $0.785 \pm 0.052$ \\
        & 8  & sae & $0.699 \pm 0.136$ & $0.759 \pm 0.076$ \\
        & 12 & raw & $0.763 \pm 0.136$ & $0.863 \pm 0.063$ \\
        & 12 & sae & $\mathbf{0.817 \pm 0.102}$ & $\mathbf{0.877 \pm 0.025}$ \\
      \cmidrule(lr){1-5}
      \multirow{4}{*}{RF3}
        & 12 & raw & $0.738 \pm 0.095$ & $0.777 \pm 0.105$ \\
        & 12 & sae & $0.706 \pm 0.076$ & $0.771 \pm 0.075$ \\
        & 16 & raw & $\mathbf{0.776 \pm 0.098}$ & $\mathbf{0.776 \pm 0.058}$ \\
        & 16 & sae & $0.758 \pm 0.127$ & $0.774 \pm 0.088$ \\
      \midrule
      \multicolumn{5}{l}{\textit{ToxinPred3 ($n$=1200)}} \\
      \cmidrule(lr){1-5}
      \multirow{4}{*}{RF3}
        & 12 & raw & $0.807 \pm 0.069$ & $0.844 \pm 0.014$ \\
        & 12 & sae & $\mathbf{0.848 \pm 0.062}$ & $\mathbf{0.851 \pm 0.016}$ \\
        & 16 & raw & $0.803 \pm 0.061$ & $0.833 \pm 0.017$ \\
        & 16 & sae & $0.820 \pm 0.060$ & $0.835 \pm 0.021$ \\
      \bottomrule
    \end{tabular}
  \end{small}
  \vspace{-3ex}
\end{table}

\begin{figure}[tb]
\centering
\includegraphics[width=\linewidth]{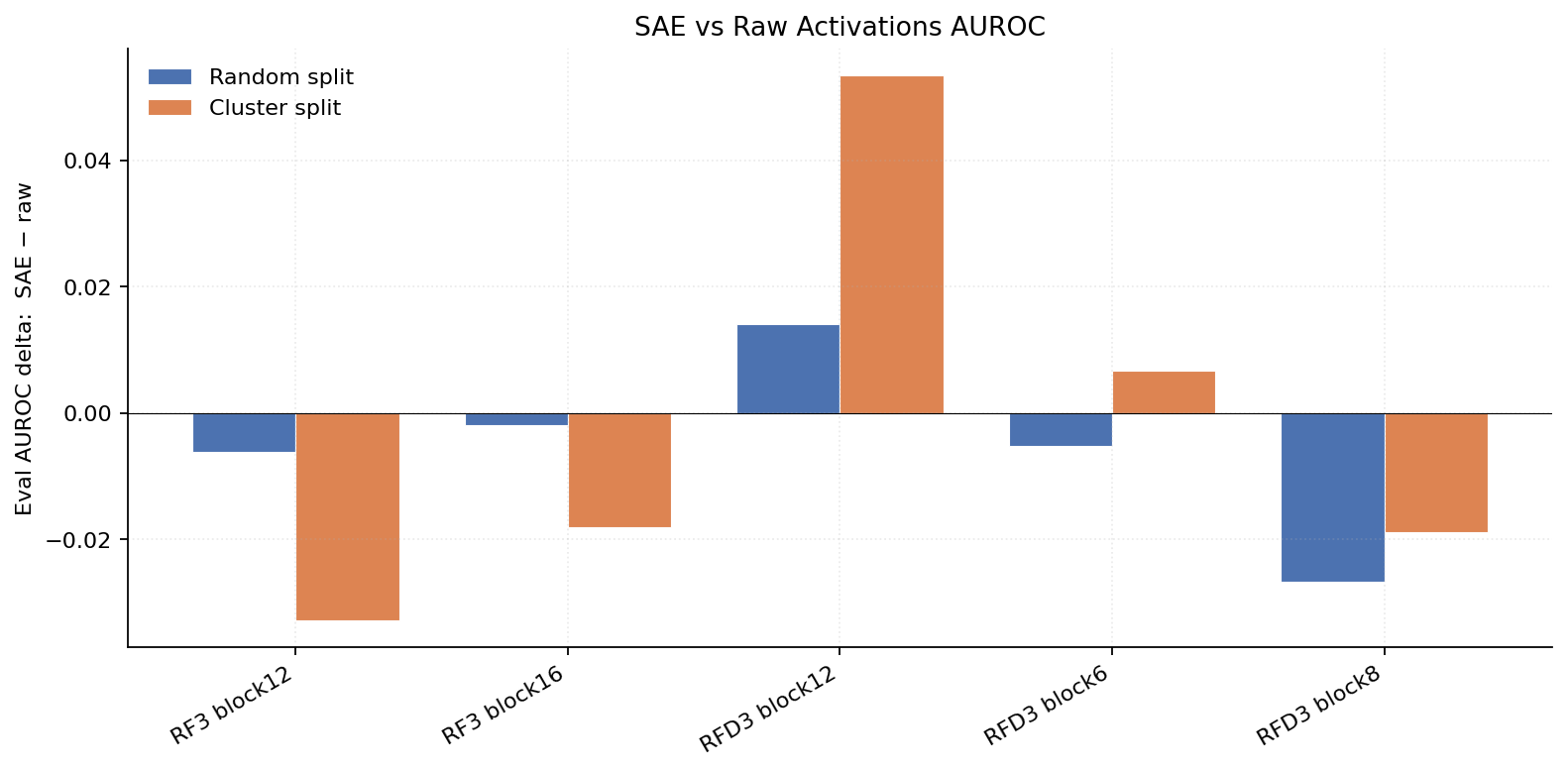}
\caption{AUROC gap (SAE $-$ raw) per block and split. The only consistent positive spike is at RFD3 block 12, cluster split ($+0.054$).}
\label{fig:sae_vs_raw}
\vspace{-2ex}
\end{figure}

\begin{figure}[tb]
\centering
\includegraphics[width=\linewidth]{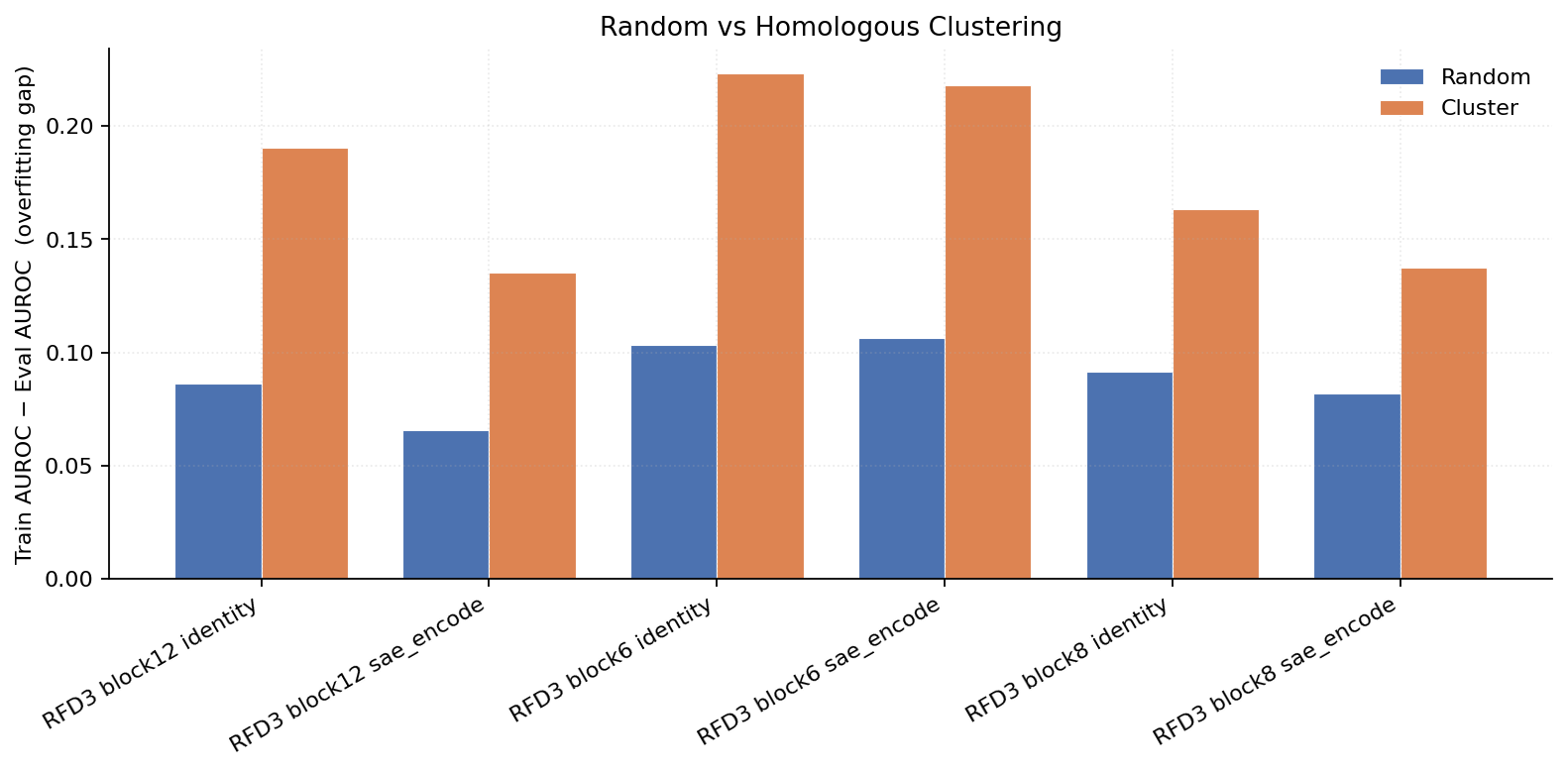}
\caption{Random$-$homology AUROC gap. Larger = more family memorization. Block 6 peaks; block-12 SAE is smallest.}
\vspace{-2ex}
\label{fig:gap}
\end{figure}

\begin{figure*}[tb]
\centering
\includegraphics[width=0.95\linewidth]{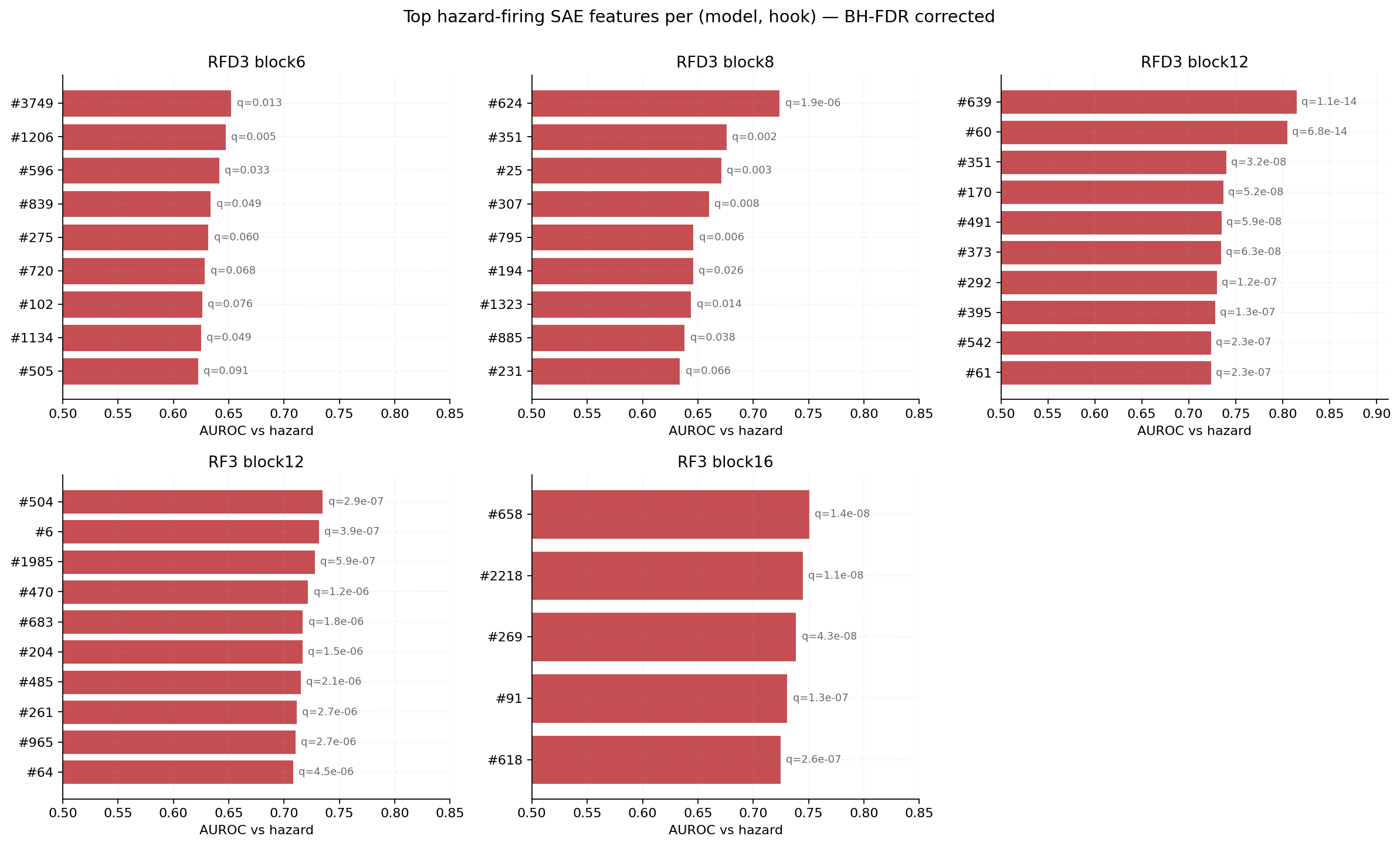}
\vspace{-1ex}
\caption{Top-10 hazard-firing SAE features per (model, block), ranked by univariate AUROC with BH-corrected $q$-values. Feature quality (peak AUROC and number of discoveries) grows with depth in RFD3, with block-12 features reaching AUROC ${\geq}0.80$.}
\label{fig:top_features}
\vspace{-2ex}
\end{figure*}

\section{Results}
\label{sec:results}

\Cref{tab:probes} reports held-out AUROC across all conditions. The best overall probe is RFD3 block-12 SAE features: AUROC $0.877 \pm 0.025$ random, $0.817 \pm 0.10$ clustered. Performance grows with depth in RFD3 (block 6 clustered: $0.592$; block 12 clustered: $0.817$) but is flat across RF3 blocks.

\subsection{SAE features beat raw activations at block 12}

\Cref{fig:sae_vs_raw} shows the AUROC delta (SAE minus raw) per block. The SAE adds $+0.054$ at RFD3 block 12 under cluster splits and is neutral or negative everywhere else. For RF3 / ToxinPred3, the SAE also adds $+0.041$ at block 12 under cluster splits. We hypothesize that the block-12 gain is due to polysemanticity: by that depth, the residual stream mixes many structural concepts, and dictionary learning untangles them into directions that a low-capacity probe can use.

The features fire on proteins from distinct hazard classes: a viper presynaptic neurotoxin, a Monkeypox immune-evasion phosphatase, a plant ribosome-inactivating protein, and a mosquito salivary toxin (See \Cref{fig:features}). Activations are localized, with each feature highlighting a handful of residues per structure. Feature 639 is the most spatially coherent, with its top three residues (99, 104, 109) all falling within a single annotated alpha-helix (positions 96--114) of ammodytoxin A. This pattern reproduces across independent partial-diffusion runs (\cref{app:moreviz}).

\subsection{Homology memorization audit}

The random-homology AUROC gap (\cref{fig:gap}) is largest at RFD3 block 6 ($0.22$ cluster), consistent with early denoising blocks collapsing toward fold-family templates. The block-12 SAE has the smallest gap ($0.13$), suggesting this is where class-discriminative, family-generalizable structure lives. RFD3's earlier blocks underperform under homology-clustering while RF3's do not, suggesting that RFD3 develops more complex, non-homology-driven features at later blocks.

\subsection{Individual hazard features}

After BH correction, significant features ($q < 0.05$) grow from ${\sim}70$ at RFD3 block 6 to ${\sim}340$ at block 12, where the top feature reaches AUROC $0.819$ ($q \approx 10^{-13}$). \Cref{fig:top_features} shows the top-10 features per (model, block) ranked by univariate AUROC, illustrating both the depth trend and the range of per-feature discrimination. \Cref{fig:features} maps activations for the four leading features onto diffusion output structures.

\subsection{Comparison with sequence-only classifiers}

Our best probe (RFD3 block-12 SAE, random, $0.877$) is competitive with sequence-only specialists: VF-Pred \cite{singh2024vfpred} ($0.84$), VirulentPred \cite{garg2008virulentpred} ($0.86$), and DTVF \cite{sun2024dtvf} ($0.92$). This is not a fair head-to-head comparison: we train a simple logistic probe rather than a task-specific model, and our dataset ($n = 275$) differs from the standard benchmark. The point is that model-internal probing can approach specialist accuracy while also localizing the signal to individual residues.

\section{Discussion}
\label{sec:discussion}

\paragraph{Block selection}
\citet{gao2025scaling} and \citet{templeton2024scaling} report that SAE feature quality peaks at middle-to-late transformer layers; block 12 of 18 in RFD3 is the analogous position. All three of our block-12 metrics converge: largest SAE-over-raw gain (\cref{fig:sae_vs_raw}), smallest memorization gap (\cref{fig:gap}), and most significant features. 

\paragraph{Limitations}
The dataset is small ($n = 275$, ${\sim}55$ per fold), so confidence intervals are wide. Hookpoint selection followed LLM practice rather than a principled ablation. Mean-pooling discards spatial information that a per-residue probe could use. Preliminary activation-steering experiments showed no effect on DTVF-scored outputs (\cref{app:steering}); steering in the partial-diffusion setting remains an open problem.

\section{Conclusion}

We showed that an SAE trained on RFdiffusion3's block-12 residual stream improves virulence probe AUROC over raw activations ($+0.054$ under homology-clustered splits), and that individual SAE features fire selectively on hazardous structures with per-residue localization. This opens a path to runtime monitoring and interpretable auditing of generative protein design models.

\bibliography{example_paper}
\bibliographystyle{icml2026}

\appendix
\onecolumn

\section{SAE Training Hyperparameters}
\label{app:hparams}

\begin{table}[h]
\centering
\begin{tabular}{ll}
\toprule
Activation dim $d$ & 768 \\
Dictionary size $m$ & 12{,}288 ($16\times$) \\
Top-K & 80 \\
Group fractions & $(0.0625, 0.125, 0.1875, 0.25, 0.375)$ \\
aux-K alpha & $0.03125$ \\
Learning rate & $2.3 \times 10^{-4}$ \\
Warmup / total steps & $1000$ / $20{,}000$ \\
Batch size & $4096$ \\
Hardware & 1 $\times$ L40 (48\,GB) \\
\bottomrule
\end{tabular}
\caption{SAE training hyperparameters (all (model, block) pairs identical).}
\end{table}

\section{Per-Fold Probe Results}
\label{app:perfold}

\begin{table}[h]
\centering
\begin{tabular}{lcc}
\toprule
Fold & N(eval) & AUROC \\
\midrule
0 & 59 & 0.982 \\
1 & 54 & 0.770 \\
2 & 54 & 0.777 \\
3 & 54 & 0.717 \\
4 & 54 & 0.838 \\
\midrule
Mean $\pm$ std & --- & $0.817 \pm 0.102$ \\
\bottomrule
\end{tabular}
\caption{Per-fold AUROC, RFD3 block 12, SAE, homology-clustered splits.}
\label{tab:perfold_block12}
\end{table}

\section{Additional Feature Visualizations}
\label{app:moreviz}

\begin{figure}[h]
\centering
\begin{subfigure}[t]{0.31\linewidth}
  \includegraphics[width=\linewidth]{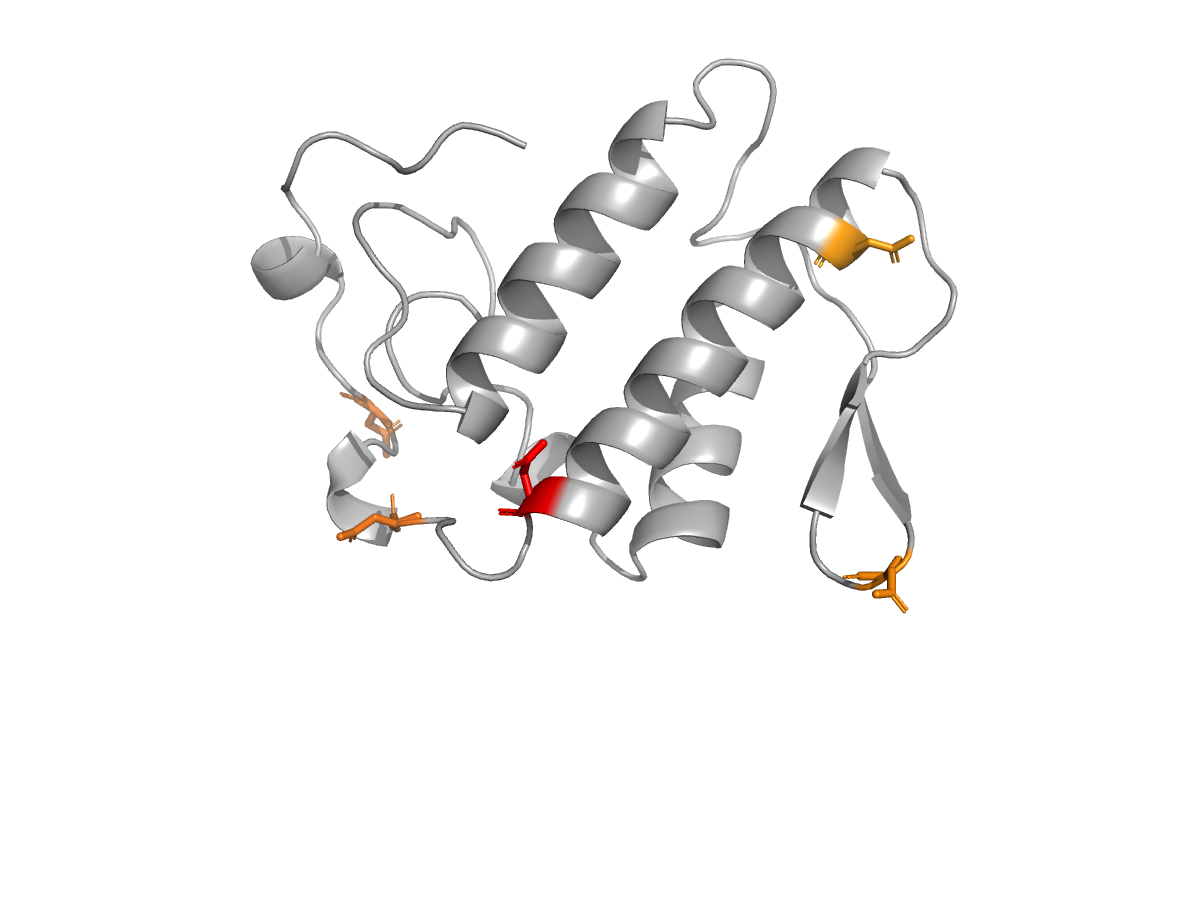}
  \caption{Feature 639, ammodytoxin A, run 2. Residues 99, 104, 109 highlighted again.}
\end{subfigure}\hfill
\begin{subfigure}[t]{0.31\linewidth}
  \includegraphics[width=\linewidth]{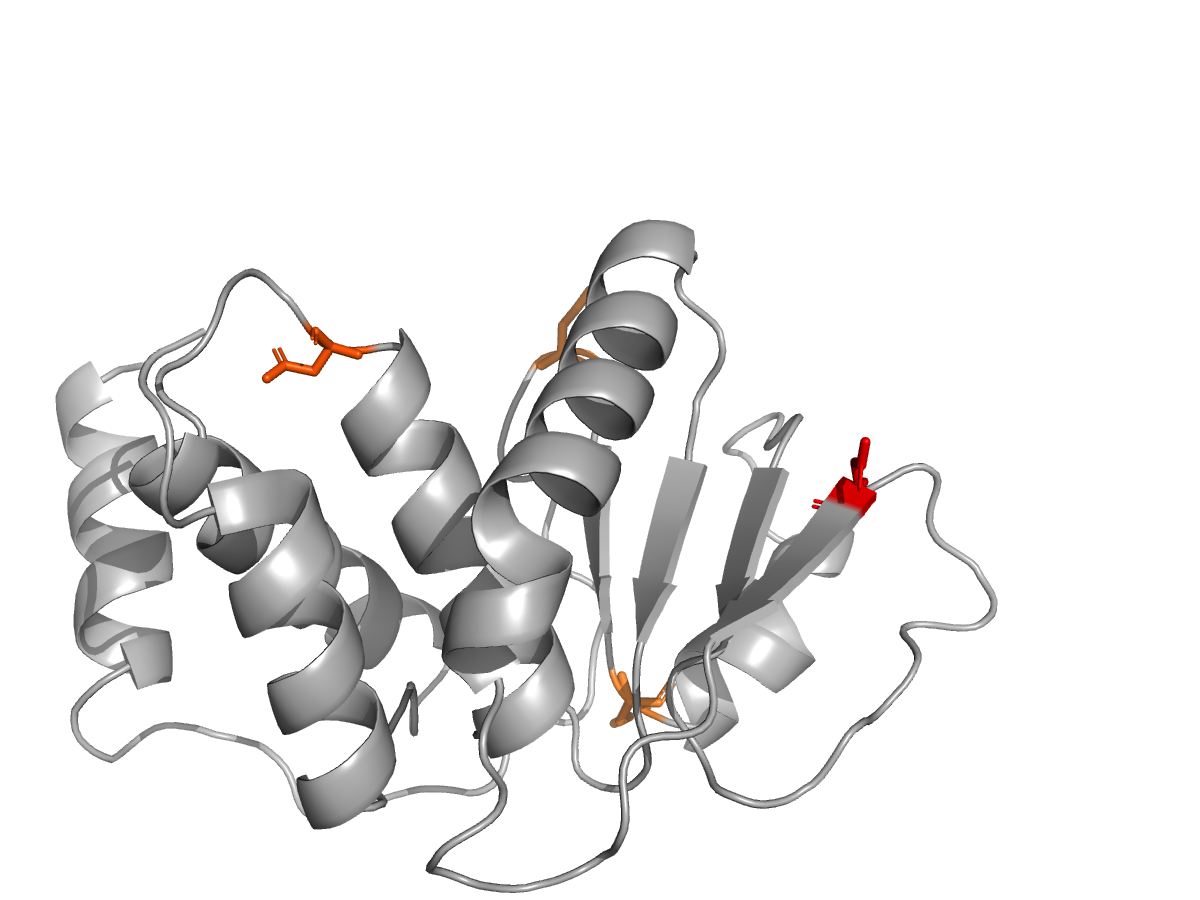}
  \caption{Feature 60, MPXV OPG106, run 2.}
\end{subfigure}\hfill
\begin{subfigure}[t]{0.31\linewidth}
  \includegraphics[width=\linewidth]{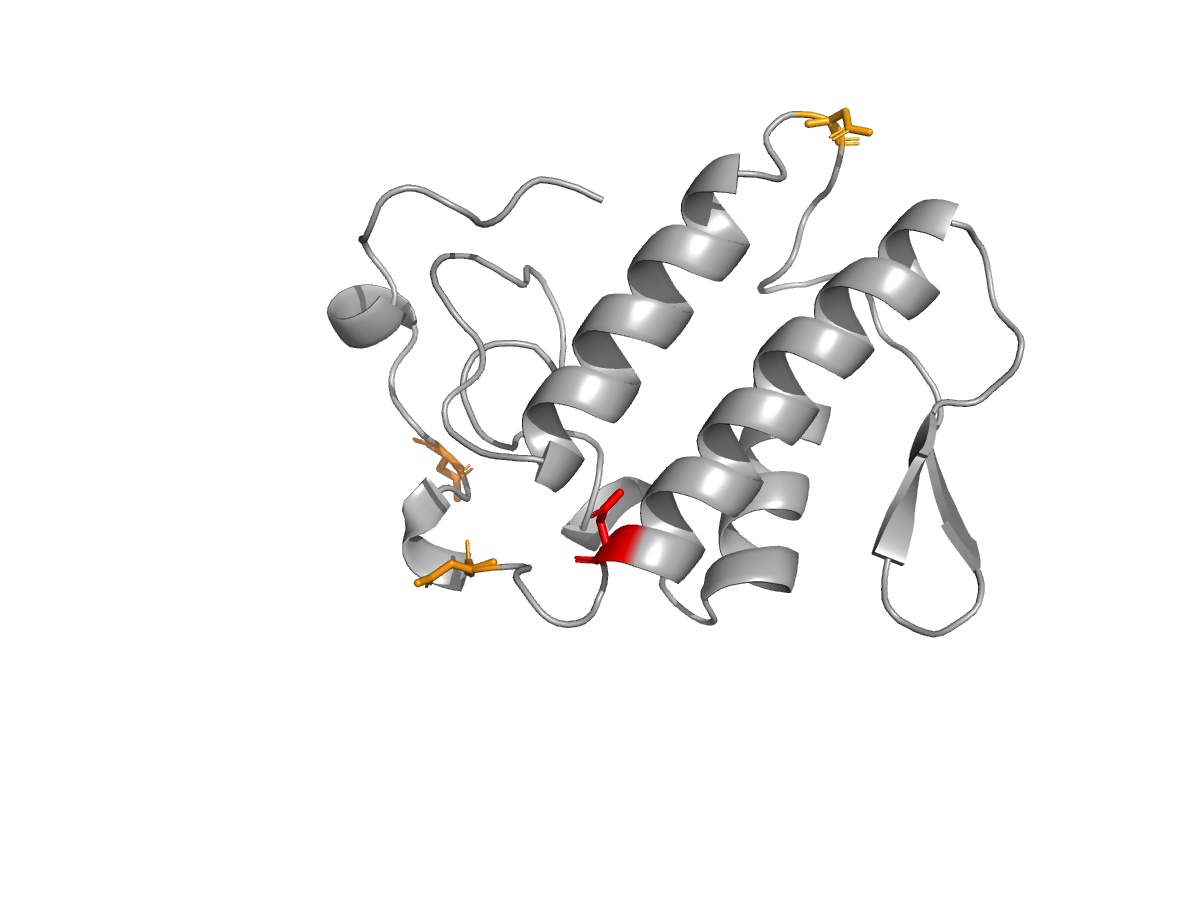}
  \caption{Feature 491 (AUROC 0.743), ammodytoxin A. Different residues than feature 639, suggesting distinct subunits.}
\end{subfigure}
\caption{Reproducibility check and additional feature. Same color scale as \cref{fig:features}.}
\label{fig:feat_extra}
\end{figure}

\section{Negative Result: Steering}
\label{app:steering}

We added a diff-of-means hazard direction (block 12, $\alpha \in \{1,2,4,8\}$) during partial-diffusion redesign of 10 hazardous PDB inputs at $\partial t \in \{5, 50\}$, then scored the designed sequences with DTVF \cite{sun2024dtvf}. DTVF probabilities were constant across $\alpha$, suggesting either the partial-diffusion setting does not provide enough denoising steps for the perturbation to propagate, or sequence extraction dominates the structural steering. We report this as a baseline for future work.

\end{document}